\documentclass[9pt,twocolumn,technote,twoside]{IEEEtran}
\usepackage{graphicx}
\usepackage{math}
\usepackage{balance}

\begin{document}

\bibliographystyle{IEEEtran}
\title{Simultaneous Robot-World and Hand-Eye Calibration}

\author{Fadi Dornaika
and Radu Horaud\IEEEcompsocitemizethanks{
Address: Inria Grenoble, 655, avenue de l'Europe, 38330 Montbonnot Saint-Martin, France. E-mail: Radu.Horaud@inria.fr}
}
\maketitle

\begin{abstract} 
Recently, Zhuang, Roth, \& Sudhakar \cite{Zhuang94} proposed a method that allows simultaneous computation of the rigid transformations from world frame to robot base frame and from hand frame to camera frame. Their method attempts to
solve a homogeneous matrix equation of the form 
$\Amat\Xmat=\Zmat\Bmat$. They use quaternions to derive explicit linear solutions for $\Xmat$ and $\Zmat$. In this short paper, we present two new solutions that attempt to solve the homogeneous matrix equation mentioned above: 
(i)~a closed-form method which uses quaternion algebra and a positive quadratic error function associated with this representation and 
(ii) a method based on non-linear constrained minimization and which simultaneously solves 
for rotations and translations. These results may be useful to other problems that can be formulated in the
same mathematical form.
We perform a sensitivity analysis for both our two methods and the
linear method developed by Zhuang et al. \cite{Zhuang94}. This
analysis allows the comparison of the three methods. In the light of this
comparison the non-linear optimization method, which solves for rotations and
translations simultaneously, seems to be the most stable one with respect
to noise and to measurement errors. 
\end{abstract}


\section{Introduction}

In order to use a gripper-mounted sensor (such as a camera) for a robot task, the position and orientation of the sensor frame
with respect to the gripper frame must be known. The problem of determining this
relationship is referred to as the hand-eye calibration problem. One can find this relationship by moving the robot and observing the resulting motion of the sensor. This calibration problem yields a homogeneous matrix equation of the form $\Amat\Xmat=\Xmat\Bmat$. Several closed-form solutions were proposed in the past
to solve for $\Xmat$ \cite{TsaiLenz89}, \cite{Wang92}, \cite{Park94}, \cite{ShiuAhmad89} as well as a non-linear optimization method \cite{HoraudDornaika95}. 

Recently, Zhuang et al. \cite{Zhuang94} proposed a method that allows the
simultaneous estimation of both
the transformations from the world-centered frame
to the robot-base frame and from the gripper frame to camera frame. The identification problem is cast into the problem of solving a system of homogeneous matrix equations of the form $\Amat\Xmat=\Zmat\Bmat$, 
where $\Xmat$ is the gripper-to-camera rigid transformation and $\Zmat$ is
the robot-to-world rigid transformation. Quaternion algebra is applied to derive explicit linear solutions for $\Xmat$ and $\Zmat$. 

\begin{figure}[t!]
\centering
\includegraphics[width = 0.45\textwidth]{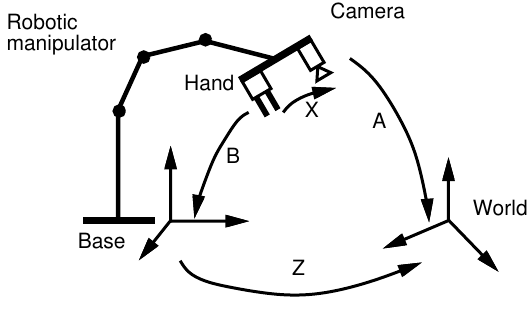}
\caption{Robot/world ($\Zmat$) and hand/eye ($\Xmat$) calibration. The camera is mounted onto the gripper and camera motions are determined using a calibration pattern. The world frame is the frame of the calibration pattern. }
\label{fig:AXYB}
\end{figure}

\begin{figure}[t!]
\centering
\includegraphics[width = 0.45\textwidth]{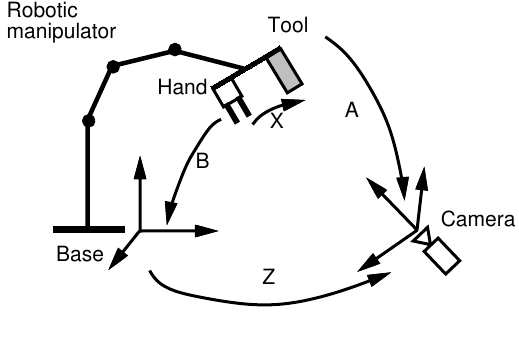}
\caption{Robot/eye ($\Zmat$) and hand/tool ($\Xmat$) calibration. The tool is mounted onto the gripper and tool motions are determined by observing tool feature points with a camera. The world frame is, in this case, identical with the camera frame.}
\label{fig:AXYB.2}
\end{figure}

The mathematical framework of $\Amat\Xmat=\Zmat\Bmat$ allows one to solve for at least two types of robotic configurations. These configurations are shown on Figure~\ref{fig:AXYB} and Figure~\ref{fig:AXYB.2}.
It is worthwhile to notice that matrices $\Xmat$ and $\Zmat$ can be estimated either
sequentially or simultaneously. Therefore two approaches are possible: 
\begin{enumerate}
\item $\Xmat$ is estimated first using any hand-eye (or camera-gripper)
calibration method and $\Zmat$ is estimated by solving the equation 
$\Amat\Xmat=\Zmat\Bmat$, or 
\item $\Xmat$ and $\Zmat$ are simultaneously estimated by solving 
$\Amat\Xmat=\Zmat\Bmat$ where both
$\Xmat$ and $\Zmat$ are unknowns.
\end{enumerate}


This paper describes both a closed-form solution and a non-linear solution for the system of matrix equations $\Amat\Xmat=\Zmat\Bmat$. These solutions solve for two
rotations and two translations that are associated with the matrices $\Xmat$ and $\Zmat$. 
Likewise the linear method \cite{Zhuang94} the closed-form and non-linear
methods yield a unique solution provided that the robot performs two motions
with distinct rotation axes. The main differences between the linear method and the
closed-form method introduced in this paper are the followings:
\begin{itemize}
\item The linear method first solves linearly for the components of two quaternions and second
it normalizes these quaternions such that they represent rotations. 
The closed-form method solves
directly for two unit quaternions and hence the constraint that these quaternions must
represent two rotations is built in the resolution method.
\item The linear method is not feasible for
some special configurations (see \cite{Zhuang94} and below). We show that the
closed-form method remains feasible for such special configurations. 
\end{itemize}

We perform a sensitivity analysis for both our methods and for the
linear method of Zhuang et al. \cite{Zhuang94}. This analysis allows the comparison of the three methods. In the light of both simulated and real experiments, 
it appears that 
the non-linear optimization method, which solves for rotations and translations simultaneously, performs better than the closed-form method which in turn
performs slightly better than the linear method. 

The remainder of this paper is organized as follows. Section~\ref{sec:problem} briefly recalls the problem formulation and presents the linear
solution suggested by Zhuang et al.
\cite{Zhuang94}. The closed-form and non-linear 
methods are described in Section~\ref{sec:solution}.
Section~\ref{sec:sensitivity} compares the three methods 
through a sensitivity analysis. Finally, Section~\ref{sec:experimental} describes 
some experimental results and Section~\ref{sec:discussion} provides a short discussion. 

\section{Problem formulation}

\label{sec:problem}

We consider an arbitrary position of the robotic system (refer to Figures \ref{fig:AXYB} and \ref{fig:AXYB.2}). From these figures we can write:
\begin{equation}
\Amat\Xmat = \Zmat\Bmat
\label{eq:AXZB}
\end{equation}

In the particular case of a camera, the matrix $\Amat$ is obtained by calibrating the camera with respect to a
fixed calibrating object and its associated frame, called the
{\em calibration frame}. The matrix $\Bmat$ is computed using the manipulator's 
forward kinematics whose parameters are supposed to be known (see
\cite{ZhuangWangRoth95} for an approach which attempts to estimate
simultaneously these kinematic parameters and the hand-eye transformation).  
Let $\Rmat_A$, $\Rmat_B$, $\Rmat_X$, and $\Rmat_Z$ be the respective 3$\times$3 rotation matrices of $\Amat$, $\Bmat$, $\Xmat$, and $\Zmat$, and let $\tvect_A$, $\tvect_B$, $\tvect_X$, and $\tvect_Z$ be the respective 3$\times$1 translational vectors. Equation (\ref{eq:AXZB}) can then be written as: 
\[
\left[
\begin{array}{cc}
   \Rmat_{A}    &  \tvect_{A} \\
   \zerovect^T &1
\end{array}
\right] \left[
\begin{array}{cc}
   \Rmat_{X}    &  \tvect_{X} \\
   \zerovect^T &1
\end{array}
\right]  =  \left[
\begin{array}{cc}
   \Rmat_{Z}  &  \tvect_{Z} \\
   \zerovect^T &1
\end{array}
\right] 
 \left[
\begin{array}{cc}
   \Rmat_{B}  &  \tvect_{B} \\
   \zerovect^T &1
\end{array}
\right]
\]
and one may easily decompose this equation into a rotation equation and a position equation:
\begin{eqnarray}
\Rmat_A\Rmat_X & = &\Rmat_Z \Rmat_B \label{eq:AXZB.rotation} \\
\Rmat_A \,\tvect_X + \tvect_A &=& \Rmat_Z\, \tvect_B + \tvect_Z
 \label{eq:AXZB.translation}
\end{eqnarray}
Equation (\ref{eq:AXZB.translation}) is a linear equation in $\tvect_X$ and $\tvect_Z$ if $\Rmat_Z$ is known. 

\subsection{Linear solution}
\label{section:linear-solution}

This solution was suggested in \cite{Zhuang94}.
Let $\qquat_A$, $\qquat_B$, $\qquat_X$, and $\qquat_Z$ be unit quaternions that correspond to the rotation matrices $\Rmat_A$, $\Rmat_B$, $\Rmat_X$, and $\Rmat_Z$
\cite{WalkerShaoVolz92}. Since quaternions can be written as a combination of a scalar and a 3-vector, we have $\qquat_A^T = \left[ a_0, \avect^T \right ]$ and so forth.
The matrix equation $\Rmat_A \Rmat_X = \Rmat_Z \Rmat_B$ is equivalent to the following quaternion equation:
\begin{equation}
\qquat_A \ast \qquat_X = \qquat_Z \ast \qquat_B
\label{eq:qaqx}
\end{equation}
Expanding eq.~(\ref{eq:qaqx}) using quaternion products yields two constraints: 
a scalar equation and a vector equation: 
\begin{eqnarray}
a_0x_0 -\avect \cdot \xvect & = & z_0b_0 - \bvect \cdot \zvect  
\label{eq:quat.scalaire}  \\
a_0\, \xvect + x_0\, \avect + \avect \times \xvect & = & z_0 \, \bvect + b_0\, \zvect - \bvect \times \zvect 
\label{eq:quat.vecteur}
\end{eqnarray}
where $\cdot$ and $\times$ denote the dot-product and the vector product in the space of 3-vectors.

If $a_0 \neq 0$, $x_0$ can be solved from (\ref{eq:quat.scalaire}):
\begin{equation}
x_0 = (\avect/a_0) \cdot \xvect + (b_0/a_0) \,z_0 - (\bvect/a_0) \cdot \zvect
\label{eq:x_0.quaternion}
\end{equation}
By substitution of eq.~(\ref{eq:x_0.quaternion}) into 
eq.~(\ref{eq:quat.vecteur}) and using the matrix representation to describe the vector and dot products yields:
\begin{eqnarray*}
& (a_0 \Imat +  \avect  \,\avect^T/a_0 + \Omega(\avect))\: \xvect + (-b_0 I - \avect  \,\bvect^T/a_0 + \Omega(\bvect) ) \:\zvect \\
& = z_0 \, \bvect - z_0\, (b_0/a_0) \, \avect 
\label{eq:a0I+aa}
\end{eqnarray*}
where $\Omega(\avect)$ is the skew-symmetric matrix associated with the 3-vector $\avect$. 

Therefore, we obtain (with $z_0 \neq 0$):
\begin{equation}
 \underbrace{\Jmat}_{3\times6} \, \underbrace{\uvect}_{6\times1} =  z_0
\left( \bvect - (b_0/a_0)  \,\avect \right)
\label{eq:Ju}
\end{equation}
where $\uvect^T = \left[ \xvect^T,\;\zvect^T\right]$.

Equation (\ref{eq:Ju}) consists of three linear constraints with six unknowns. 
Therefore, a unique solution for $\uvect$ requires multiple measurements.

The solution of $\uvect$ can be obtained using standard linear algebra techniques. After $\uvect$ is obtained, the components of both $\qquat_X$ and $\qquat_Z$ can be determined using the constraints $\| \qquat_X \| ^{2} = \| \qquat_Z\| ^{2} = 1 $ and eq.~(\ref{eq:x_0.quaternion}).

Following the solution of $\Rmat_X$ and $\Rmat_Z$, the computation of $\tvect_X$ and $\tvect_Z$ becomes trivial. Each position of the hand provides three linear equations with six unknowns (the components of $\tvect_X$ and $\tvect_Z$). 

\section{Problem solution}
\label{sec:solution}

In this section we propose two alternatives for estimating
$\Rmat_X$, $\Rmat_Z$, $\tvect_X$, and $\tvect_Z$:
A closed-form method and a non-linear method
which do not suffer from the above limitations, e.g., $a_0\neq 0$ and
$z_0\neq 0$. 

The closed-form method uses algebraic properties associated with quaternions
to cast a sum of squares error function into a positive semi-definite
quadratic form whose minimization uses two Lagrange multipliers.
The non-linear method solves for all
the unknowns simultaneously using standard minimization techniques.
Interesting enough, the closed-form method is
similar but not equivalent
to the problem of optimally estimating rigid motion from 3-D to 3-D
point or line correspondences \cite{WalkerShaoVolz92}, \cite{Faugeras93}. The method
introduced in this paper solves simultaneously for two rotations in closed form while the
methods developed in the past solved for one rotation in closed form.

\subsection{Closed-form method}
\label{section:closed-form-solution}

We start by building a positive error function that is derived from equation (\ref{eq:qaqx}) as follows.
Since the quaternion multiplication can be written in matrix form and
with the notations introduced in \cite{WalkerShaoVolz92} we have: 
\begin{eqnarray*}
\qquat_{Ai} \ast \qquat_X &=& Q(\qquat_{Ai})  \, \qquat_X \\
\qquat_Z \ast \qquat_{Bi} & = & W(\qquat_{Bi})\, \qquat_Z
\end{eqnarray*}
By substituting these equations into (\ref{eq:qaqx}), we obtain:
\[
Q(\qquat_{Ai}) \, \qquat_X - W(\qquat_{Bi})\, \qquat_Z= \zerovect
\]
With matrices $Q(\qquat)$ and $W(\qquat)$ being defined by:
\[
\begin{array}{c}
Q(\qquat) = \left[ \begin{array}{rrrr}
q_{0} & -q_{x} & -q_{y} & -q_{z} \\
q_{x} &  q_{0} & -q_{z} &  q_{y} \\
q_{y} &  q_{z} &  q_{0} & -q_{x} \\
q_{z} & -q_{y} &  q_{x} &  q_{0}
           \end{array}
	   \right]
\\
W(\qquat) = \left[ \begin{array}{rrrr}
q_{0} & -q_{x} & -q_{y} & -q_{z} \\
q_{x} &  q_{0} &  q_{z} & -q_{y} \\
q_{y} & -q_{z} &  q_{0} &  q_{x} \\
q_{z} &  q_{y} & -q_{x} &  q_{0}
           \end{array}
 	   \right]
\end{array}
\]
Moreover, these two matrices are orthogonal and for a unit
quaternion $\qquat$ we have:
\[	Q(\qquat)^TQ(\qquat) = \qquat^T\qquat\Imat = \Imat	\]
\[      W(\qquat)^TW(\qquat) = \qquat^T\qquat\Imat = \Imat       \]

The squared norm of the corresponding error vector is given by the following positive
quadratic form:
\begin{eqnarray*}
&&\| Q(\qquat_{Ai}) \,\qquat_X - W(\qquat_{Bi})\, \qquat_Z \|^2 \\
&=& \left [ Q(\qquat_{Ai}) \, \qquat_X - W(\qquat_{Bi})\, \qquat_Z \right  ]^T \; [Q\left [ \qquat_{Ai}) \, \qquat_X - W(\qquat_{Bi})\, \qquat_Z \right ] \\
 &=& \qquat_X^T   Q(\qquat_{Ai})^T Q(\qquat_{Ai}) \qquat_X +
      \qquat_Z^T  W(\qquat_{Bi})^T W(\qquat_{Bi}) \qquat_Z -\\
  &  &  \qquat_Z^T W(\qquat_{Bi})^T Q(\qquat_{Ai}) \qquat_X \\
 &&- \qquat_X^T Q(\qquat_{Ai})^T W(\qquat_{Bi}) \qquat_Z 
\end{eqnarray*}

Let $\vvect$ be an 8-vector given by:
\[ \vvect^T = \left [\qquat_X^T , \: \qquat_Z^T \right ]
\]

Thus, we can write:
\[
\| Q(\qquat_{Ai}) \, \qquat_X - W(\qquat_{Bi})\, \qquat_Z \|^2 = \vvect^T \Smat_i\: \vvect
\]
with  $\Smat_i$ being an 8$\times$8 positive semi-definite symmetric matrix:
\begin{equation} 
\Smat_i =  \left [ \begin{array}{cc}
\Imat & \Cmat_i  \\
 \Cmat^T_i  &  \Imat
\end{array} \right ]
\label{eq:symmetric-8by8}
\end{equation}
where $\Cmat_i=-Q(\qquat_{Ai})^T W(\qquat_{Bi})$ is an orthogonal matrix of
rank equal to 4.

Finally, the error function that will allow us to compute $\qquat_X$ and $\qquat_Z$ becomes ($n$ is the number of different positions of the robot):
\begin{equation}
f(\qquat_X, \qquat_Z) 
          = \sum_{i=1}^{n} \vvect^{T}\Smat_i\, \vvect 
          = \vvect^{T} \left( \sum_{i=1}^{n} \Smat_i \right) \vvect     
          =  \vvect^{T} \Smat\, \vvect  
\label{eq:f1=qTAq}
\end{equation}
with: 
\[ \Smat= \left [ \begin{array}{cc}
n\Imat & \Cmat  \\
 \Cmat^T  &  n\Imat
\end{array} \right ]
\]
Notice that $\Cmat=\sum_{i=1}^{n}\Cmat_i$ is the sum of $n$ orthogonal
matrices. In the general case $\Cmat$ has full rank and there may be geometric
configurations for which $\Cmat$ is rank deficient. However, such geometric configurations are
very rare in practice and, without loss of generality, one may assume that $\Cmat$ has always
full rank.
The function $f(\qquat_X, \qquat_Z)$ is a positive semi-definite quadratic
form and one way to minimize it is to use two Lagrange multipliers:
\begin{eqnarray*}
 \min_{\vvect} f &=&  \min_{\qvect_X,\qvect_Z} ((\qvect_X\;\qvect_Z)^{T} \Smat (\qvect_X\;\qvect_Z)\\
 && + \lambda_1\, (1-\qvect_X^{T}\qvect_X)
+ \lambda_2\, (1-\qvect_Z^{T}\qvect_Z)) 
\end{eqnarray*}

By developing and grouping terms we obtain:
\begin{eqnarray}
f(\qvect_X,\qvect_Z) &=& (n-\lambda_1) \qvect_X^{T}\qvect_X + (n-\lambda_2)
\qvect_Z^{T}\qvect_Z \nonumber \\
&& + \qvect_X^{T} \Cmat \qvect_Z + \qvect_Z^{T} \Cmat^T
\qvect_X + \lambda_1 + \lambda_2
\label{eq:the-error-dev}
\end{eqnarray}
This function passes through a minimum when the first derivatives vanish. By
differentiating with respect to the components of $\qvect_X$ and $\qvect_Z$ we
obtain:
\begin{eqnarray}
\label{eq:der-qx}
(n-\lambda_1) \qvect_X + \Cmat \qvect_Z & = & 0 \\
\label{eq:der-qz}
(n-\lambda_2) \qvect_Z + \Cmat^T \qvect_X  & = & 0
\end{eqnarray}
From equation~(\ref{eq:der-qx}) we obtain:
\begin{equation}
\qvect_X = \frac{1}{\lambda_1 - n} \Cmat \qvect_Z
\label{eq:sol-qx}
\end{equation}
and by substituting $\qvect_X$ in equation~(\ref{eq:der-qz}) we obtain:
\begin{equation}
\Cmat^T\Cmat \qvect_Z = (\lambda_1 - n)(\lambda_2 - n) \qvect_Z
\label{eq:sol-qz}
\end{equation}
Therefore $\qvect_Z$ is an eigenvector of the symmetric positive semi-definite
matrix $\Cmat^T\Cmat$. Such a matrix has four real positive eigenvalues
$\alpha_i,\;i=\{1...4\}$ and we have an eigenvector $\evect_i$ for each
eigenvalue:
\[	\Cmat^T\Cmat \evect_i = \alpha_i \evect_i	\]
Notice that by substituting equations~(\ref{eq:sol-qx}) and (\ref{eq:sol-qz})
into equation~(\ref{eq:the-error-dev}) we obtain the value of the error
function at the point where the first derivatives vanish:
\[	f(\qvect_X,\qvect_Z) = \lambda_1 + \lambda_2	\]
Therefore, we must choose an eigenvalue $\alpha_i$ which minimizes $\lambda_1
+ \lambda_2$. Let us consider the fact that $\qvect_X$ must be a unit
quaternion. We obtain from equations (\ref{eq:der-qx}) and (\ref{eq:sol-qz}):
\begin{eqnarray*}
\qvect_X^{T}\qvect_X &=& \frac{1}{(\lambda_1 - n)^2}\qvect_Z^{T} \Cmat^T
\Cmat \qvect_Z \\
&=& \frac{1}{(\lambda_1 - n)^2}\qvect_Z^{T}
(\lambda_1 - n)(\lambda_2 - n) \qvect_Z\\
& = & \frac{\lambda_2 - n}{\lambda_1 - n}
= 1
\end{eqnarray*}

Hence, we must have:
\[	 \lambda_1 = \lambda_2 \neq 0	\]
The relationship between $\lambda_1=\lambda_2=\lambda$ and $\alpha_i$, i.e.,
equation~(\ref{eq:sol-qz}) is:
\[	(\lambda - n)^2 = \alpha_i	\]
which yields the following solutions for $\lambda$:
\[	\lambda = n \pm \sqrt{\alpha_i}	\]
Since $\lambda$ must be a positive number, one has to select among the four positive
eigenvalues, the eigenvalue
$\alpha_i$ such that $n\pm\sqrt{\alpha_i}$ is the smallest positive number.
 
Once the rotations, $\Rmat_X$ and $\Rmat_Z$, have been determined, the problem
of determining the best
translations, $\tvect_X$ and $\tvect_Z$, becomes a linear least-squares
problem
that can be easily solved
using standard linear algebra techniques.

\subsubsection{Configurations defeating the linear method}
There are two configurations for which the linear method fails to provide a
solution: $z_0=0$ and $a_0=0$ 
(see Section~\ref{section:linear-solution}). Clearly the closed-form solution 
is able to deal with situations for which $z_0=0$. 
The case $a_0=0$ is a little bit more
complex to analyse.
First, notice that the
4$\times$4 matrices Q(\qquat) and W(\qquat) have full rank for all non null
quaternions $\qvect$. 
Let, for some $i$,
$\qquat_{Ai}= [0,\avect^T_i]^T$. $Q(\qquat_{Ai})$ becomes a skew-symmetric
matrix of full rank for all $\avect_i\neq
\zerovect$. Hence, the rank of $\Smat_i$ in equation~(\ref{eq:symmetric-8by8})
is not affected by such a
special case. However there is an ambiguity associated with purely imaginary
unit quaternions because the quaternions $\qquat_{Ai}= [0,\avect^T_i]^T$
and $\qquat_{Ai}= [0,-\avect^T_i]^T$ describe the same rotation matrix
$\Rmat_{Ai}$. Hence, one has two consider two distinct matrices associated
with this special configuration:
\[
\begin{array}{ccc}
\Smat_i^+ =  \left [ \begin{array}{cc}
\Imat & \Cmat_i  \\
 \Cmat^T_i  &  \Imat
\end{array} \right ]
& \mbox{and} &
\Smat_i^- =  \left [ \begin{array}{cc}
\Imat & -\Cmat_i  \\
 -\Cmat^T_i  &  \Imat
\end{array} \right ]
\end{array}
\]
Therefore, any time such a special configuration is present in the data, one
has to consider two distinct error functions. There will be two two distinct
solutions for $\qvect_X$ and $\qvect_Z$. One may simply consider, among these
two solutions, the solution
yielding the smallest minimum.

\subsection{Non-linear method}
\label{section:non-linear-solution}

There are several disadvantages associated with the above methods:
\begin{enumerate}
\item The unknowns are estimated in sequence, rotations first and
then translations. Errors from the first stage propagate to the second stage;
\item It is well known that the performance of linear resolution methods 
degrades in
the presence of noise, and 
\item Unlike non-linear minimization, linear and closed-form solutions do not
allow a characterization of both the quality of the solution and the
confidence associated with the solution.
\end{enumerate} 

In this Section, we propose to overcome the disadvantages mentioned above. For this purpose we estimate simultaneously the rotations and translations associated with $\Xmat$ and $\Zmat$.  This leads to a non-linear minimization problem. There are
24 parameters associated with two rotation matrices (18 parameters) and two translation
vectors (6 parameters). The initialization of these unknowns is straightforward
because one can use either of the two methods outlined above. Non-linear
minimization provides information about both the quality of the solution
(the depth of the global minimum) and the confidence associated with this
solution (the width of the global minimum).

If we have $n$ positions of the robot, the calibration problem becomes the problem of solving for a set of $2n$ non-linear constraints derived from equations (\ref{eq:AXZB.rotation}) and
(\ref{eq:AXZB.translation}), or equivalently, the problem of minimizing the following error function:
\begin{eqnarray*} 
 f(\Rmat_X, \Rmat_Z, \tvect_X, \tvect_Z)
 &=& \mu_1 \, \sum_{i=1}^{n}  
\left( \|\Rmat_{Ai} \, \Rmat_X - \Rmat_{Bi} \, \Rmat_Z   \| ^2 \right)\\
&+&  
\mu_2 \,  \sum_{i=1}^{n}  \left(\|\Rmat_{Ai} \,\tvect_X + \tvect_{Ai} - \Rmat_Z \,\tvect_{Bi} - \tvect_Z \| ^2 \right)   \\ 
& +&  \mu_3 \, \|\Rmat_X\Rmat_X^T - \Imat\|^2 + \mu_4 \,  \|\Rmat_Z\Rmat_Z^T - \Imat\|^2 
\end{eqnarray*}
The criterion to be minimized is of the form:
\[
\min _{\xvect} \left\{ f(\xvect)= \frac{1}{2} \sum_{j=1}^{m} \Phi_{j}^{2}(\xvect): \xvect \in I\!\!R^{24} \right\}
\]

Therefore, the problem becomes a classical non-linear least-squares 
constrained minimization 
problem and one can apply standard non-linear optimization techniques, such as Newton's method and Newton-like methods \cite{GillMurrayWright89}, \cite{Fletcher90}.  In this error function, the terms $\Phi_j$ are quadratic in the unknowns. Notice that the last two terms
are {\em penalty} functions which constrain the matrices $\Rmat_X$ and $\Rmat_Z$ to be
rotations. 
The parameters $\mu_1$ through $\mu_4$ are real positive numbers. 
High values for $\mu_3$ and $\mu_4$ inforce the role of the penalty functions
In all our experiments we have set these 
parameters to the following
values:
$\mu_1 = \mu_2 =1$ and
$\mu_3 = \mu_4 = 10 ^6$
In the next two sections we give some results obtained with the
Levenberg-Marquardt non-linear minimization method as described in \cite{NumericalRecipes} and in \cite{Fletcher90}.

\section{Sensitivity analysis and method comparison}
\label{sec:sensitivity}

One of the most important merits of any calibration method is its
sensitivity with respect to various perturbations. In our problem, there are two main sources of
perturbations: errors associated with camera calibration and errors associated
with the robot position. Indeed, the parameters of both the direct and
inverse kinematic models of robots are not perfect. In order to investigate the behaviour of the methods in the presence of measurement noise we designed a 
sensitivity analysis based on the following grounds:

\begin{itemize}
\item Nominal values for the parameters of both the hand-eye transformation $\Xmat$ and the robot-to-world transformation $\Zmat$ are provided;
\item Also are provided $n$ matrices $\Amat_1$, \ldots $\Amat_{n}$ from which $n$ 
hand positions can be computed  with:
\[      \Bmat_{i} = \Zmat^{-1} \Amat_{i} \Xmat         \]

\item Either Gaussian noise or uniform noise is added to both camera and robot
positions; the homogeneous transformations, ($\Xmat$ and $\Zmat$), are estimated in the presence of this noise using
the three methods described in this paper: the linear method, the
closed-form method and the non-linear method, and
\item We study the variations of the estimation of the hand-eye transformation
and the robot-to-world transformation as a function of the noise being added and/or as a function of the number of
positions ($n$).
\end{itemize}

Since both rotations and translations may be represented as vectors, adding
noise to a transformation consists of adding random numbers to each one of the
vectors' components. Random numbers simulating noise are obtained using a
random number generator either with a uniform distribution in the interval 
$[-C/2, +C/2]$, or with a Gaussian distribution with a standard deviation
equal to $\sigma$. Therefore
the level of noise that is added is associated either with
the value of $C$ or with the value of $\sigma$ (or, more precisely, with the
value of 2$\sigma$). In what follows the level of noise is in fact represented
as a ratio: the amplitude of the actual random numbers ($C$ or 2$\sigma$) divided by the nominal values of the perturbed parameters.

In the case of a rotation, the vector (quaternion)
associated with this rotation has a
module equal to 1 and therefore the ratio is simply either $C$ or 2$\sigma$.
In the case of a translation the ratio is computed with respect to a nominal
value estimated over all the perturbed translations:
\[
\| \tvect_{nominal} \| = 
\frac{ \sum_{i=1}^n ( \| \tvect_{A_{i}} \| + \| \tvect_{B_{i}} \| ) }{2n}
\]
where $\tvect_{A_{i}}$ is the translation vector associated with $\Amat_{i}$.

For each noise level and for a large number $N$ of trials we compute the errors as follows. These errors are: orientation error and position error. The orientation error is defined as 
the rotation angle in degrees required to align the coordinate system of $\Xmat$ or $\Zmat$ in 
its computed orientation with the coordinate system in its theoretical orientation. 
The position error is defined as the norm of the vector which represents the difference between the two 
translation vectors: the computed one and the theoretical one, divided by the norm of the second vector.  

In all our simulations we set
$N=500$, $\| \tvect_X\| = 229$mm, and $\| \tvect_Z\|=768$mm.

The following figures show the average of the above errors as a function of the percentage of
noise. The percentage of noise varies from 1\% to 6\%. The full curves (---)
correspond to the method in \cite{Zhuang94}, the dotted curves ($\ldots$) 
correspond to the closed-form method, and the dashed curves (- - -)
correspond to the non-linear method.

\begin{figure}[t!]
\centering
\includegraphics[width = 0.4\textwidth]{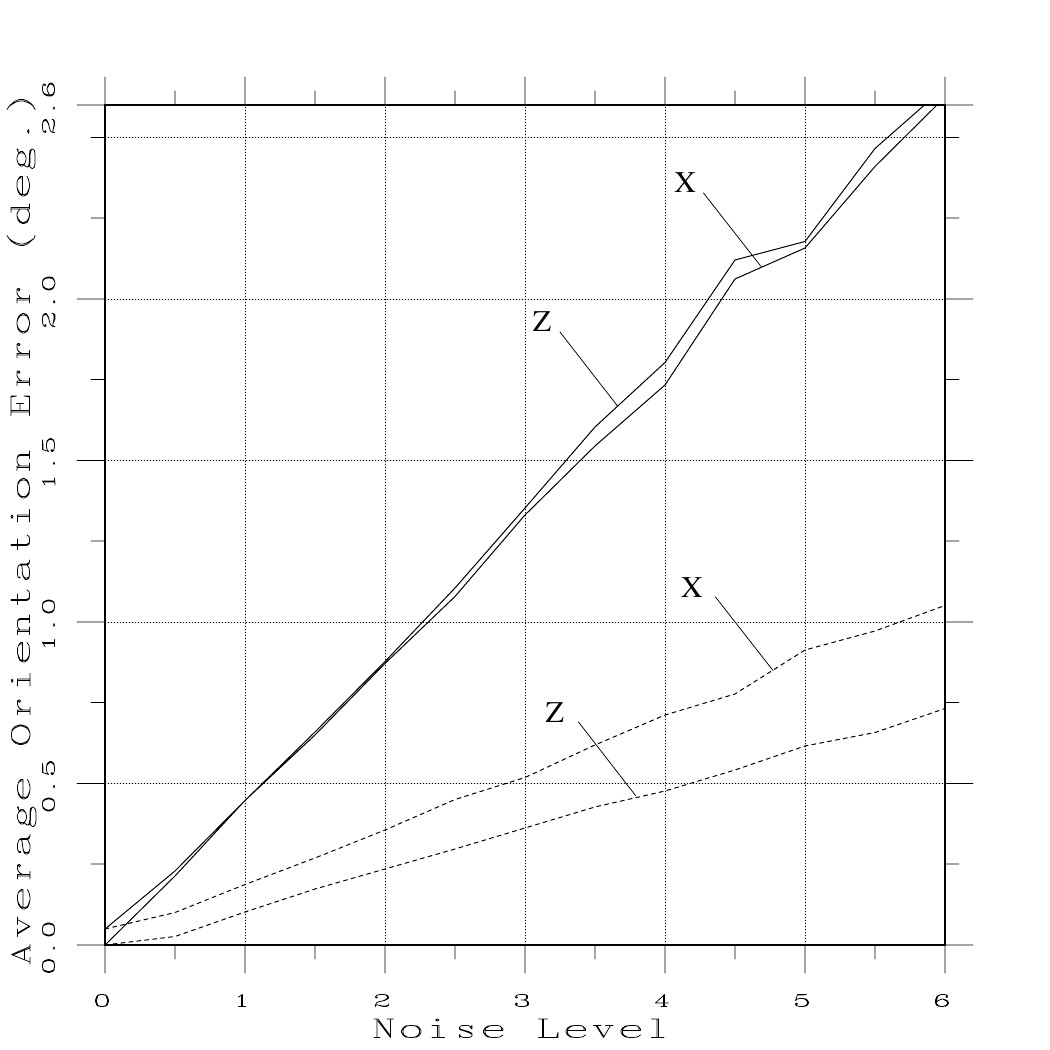}
\centerline{(a) Orientation errors.}
\includegraphics[width = 0.4\textwidth]{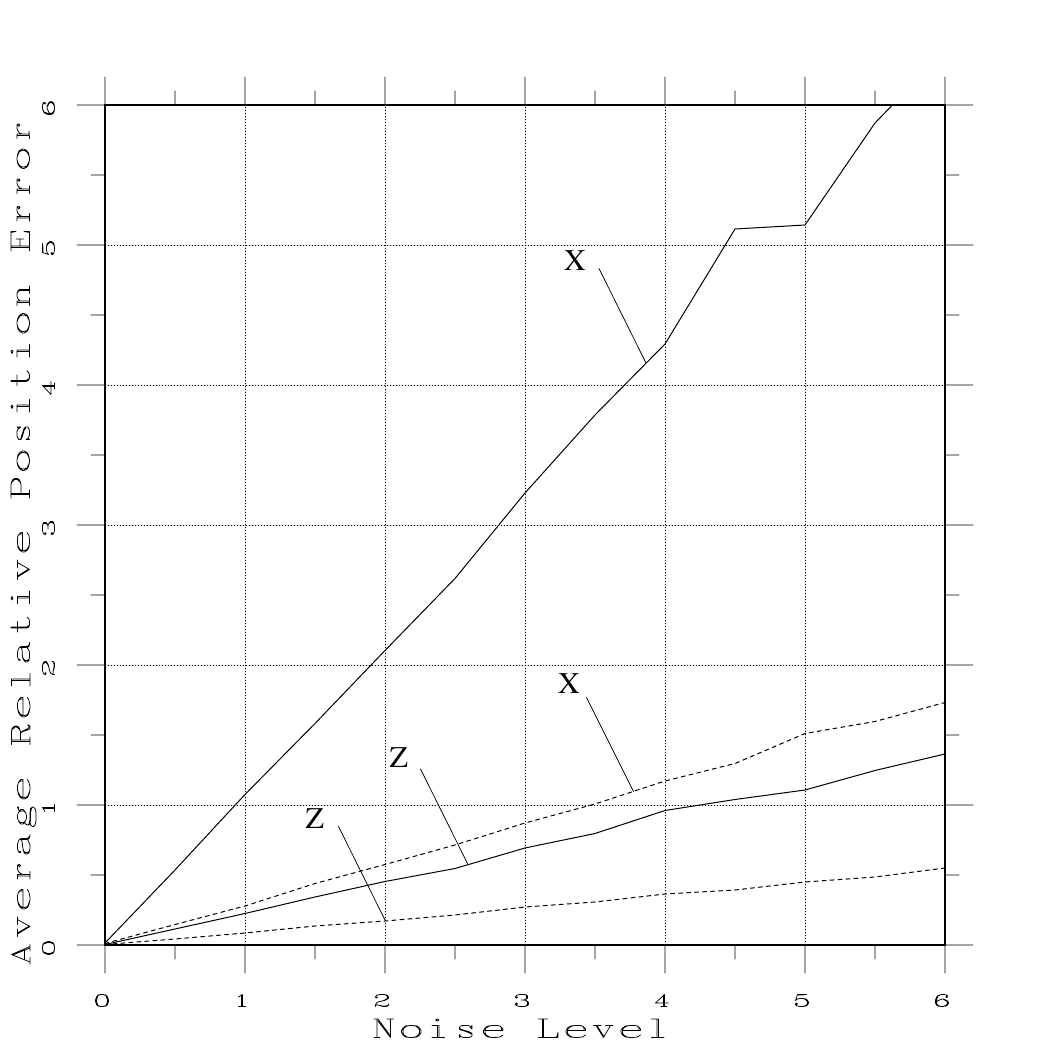}
\centerline{(b) Relative position errors.}
\caption{Errors in orientations and positions in the presence of uniform noise perturbing the rotation axes. The full curves (---) correspond to the method of \cite{Zhuang94} and the dashed curves (- - -) correspond to the non-linear method.}
\label{fig:erreurs1}
\end{figure}

\begin{figure}[t!]
\centering
\includegraphics[width = 0.4\textwidth]{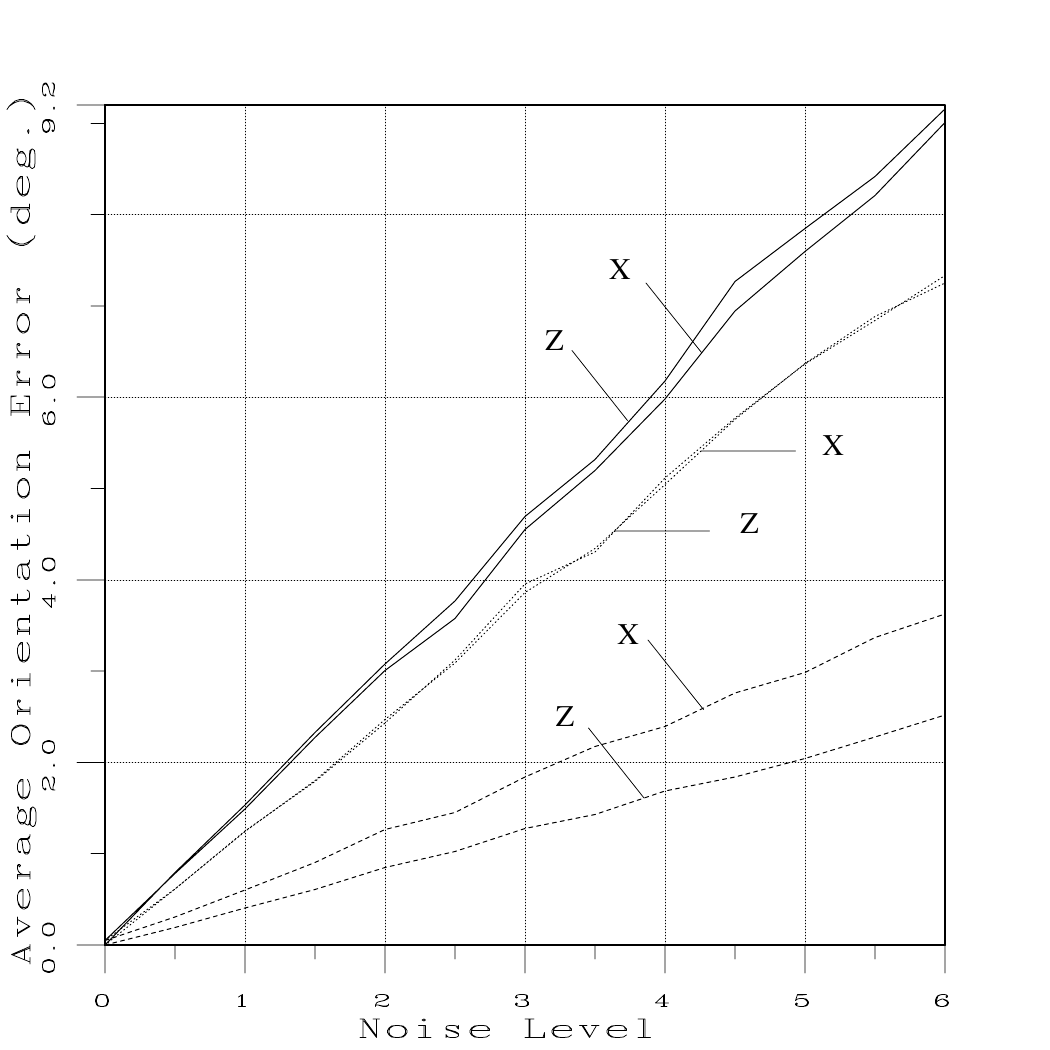}
\centerline{(a) Orientation errors.}
\includegraphics[width = 0.4\textwidth]{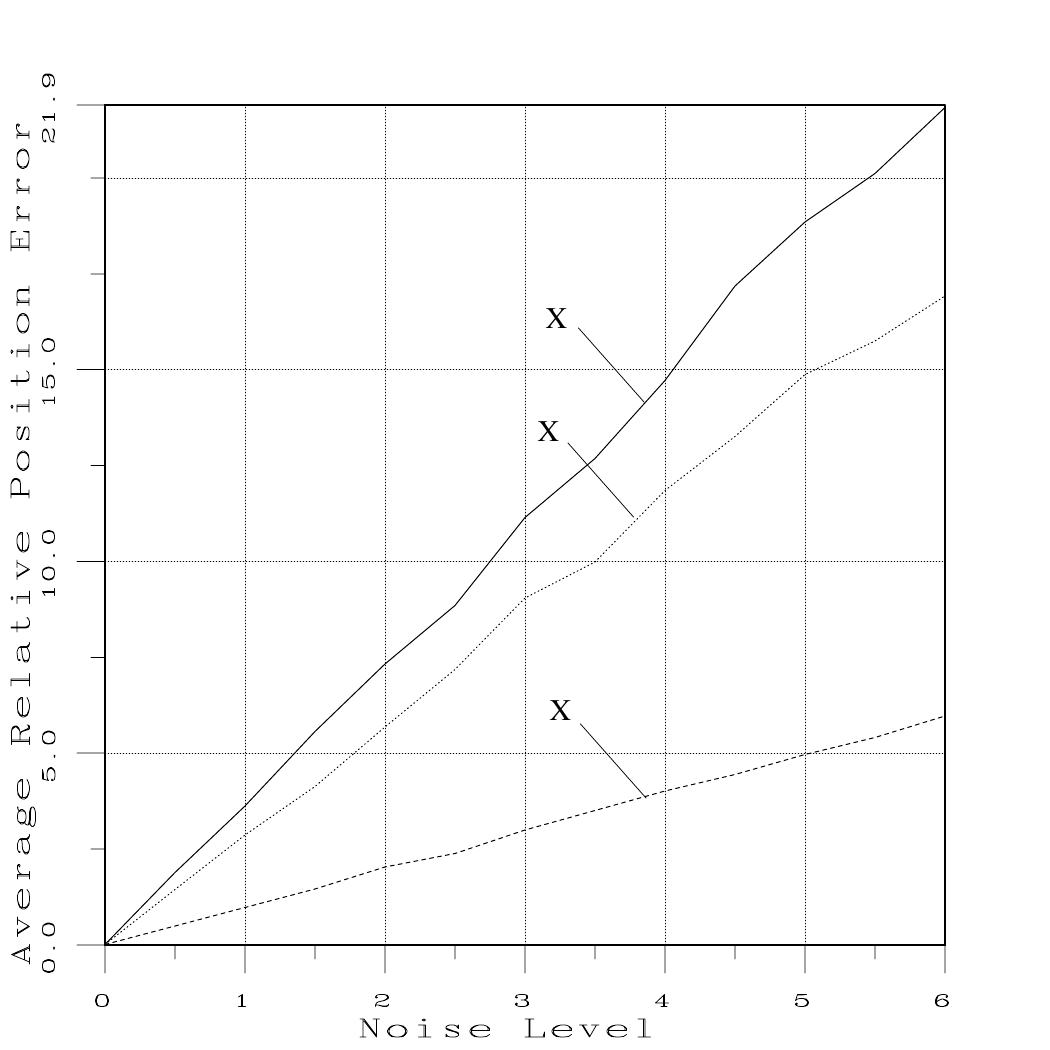}
\centerline{(b) Relative position errors.}
\caption{Errors in orientations and positions in the presence of Gaussian noise perturbing the rotation axes. The full curves (---) correspond to the method of \cite{Zhuang94}, the dotted curves ($\ldots$) correspond to the closed-form solution and the dashed curves (- - -) correspond to the non-linear method.}
\label{fig:erreurs2}
\end{figure}

\begin{figure}[t!]
\centering
\includegraphics[width = 0.4\textwidth]{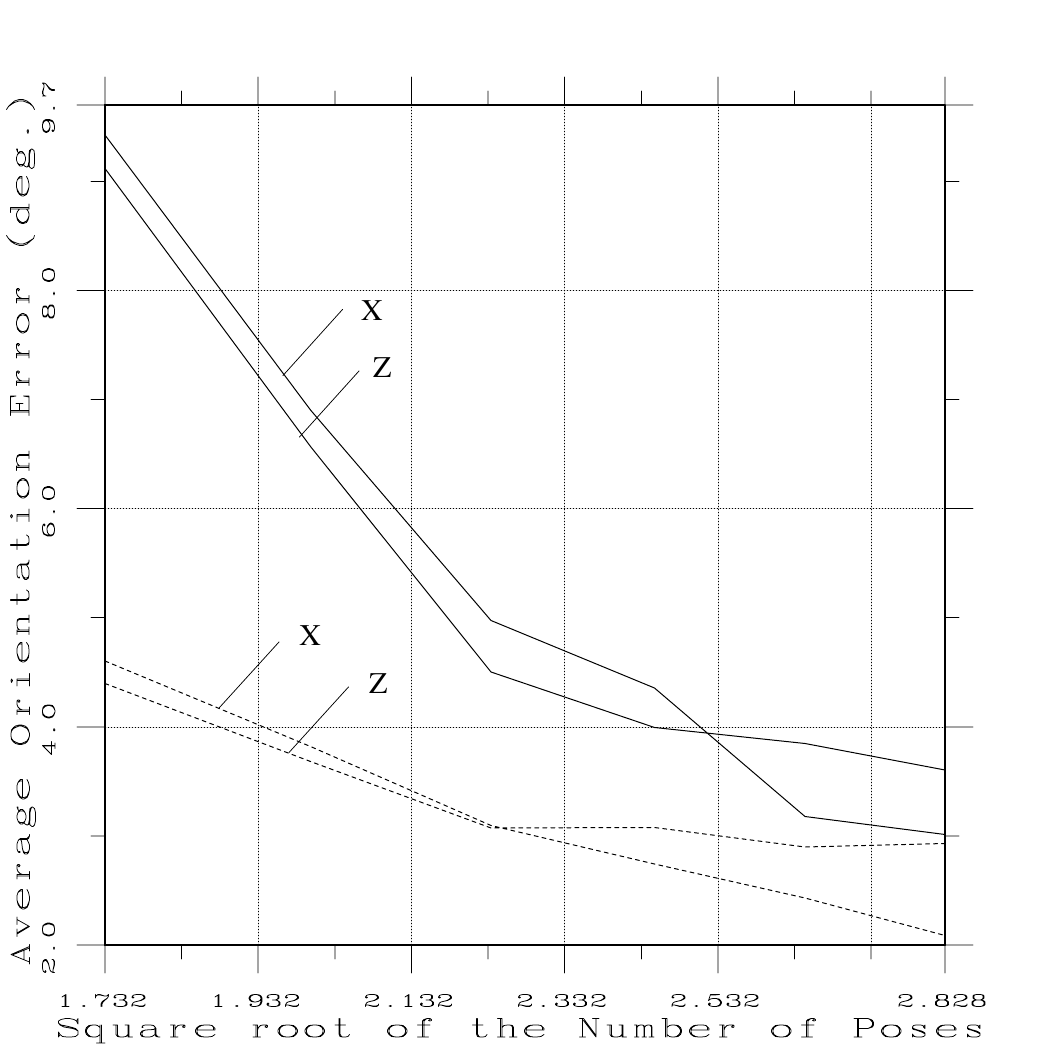}
\centerline{(a) Orientation errors.}
\includegraphics[width = 0.4\textwidth]{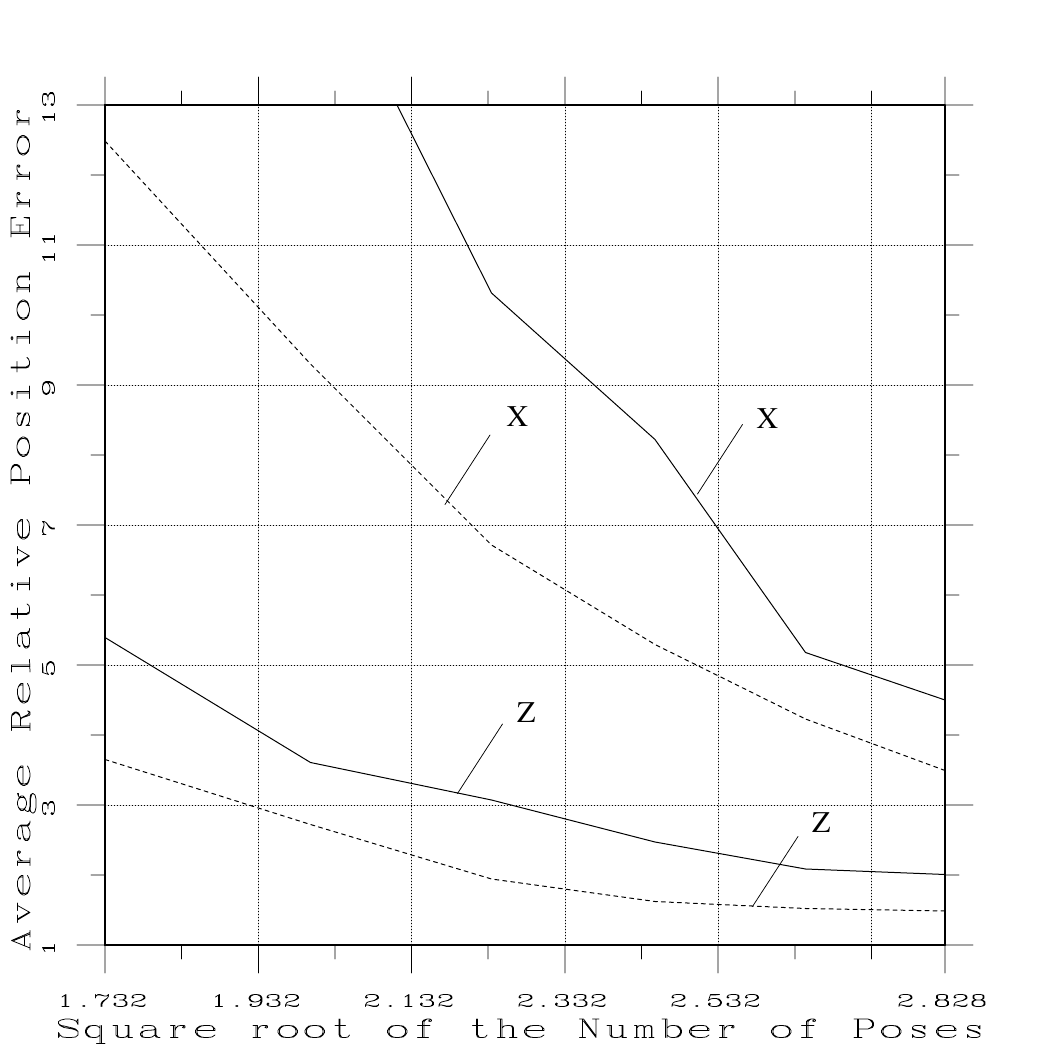}
\centerline{(b) Relative position errors.}
\caption{Errors in orientations and positions as a function of the number of positions. A Gaussian noise is added both to the robot and camera positions. The full curves (---) correspond to the method of \cite{Zhuang94} and the dashed curves (- - -) correspond to the non-linear method.}
\label{fig:erreurs4}
\end{figure}

Figures~\ref{fig:erreurs1}
and~\ref{fig:erreurs2} correspond to three positions ($n=3$) of the
robot while on Figure~\ref{fig:erreurs4} the number of positions varies from 3 to 8.

Figure~\ref{fig:erreurs1} shows the rotation and
translation errors as a function of uniform noise added to the rotational part
of the robot and camera positions. Figure~\ref{fig:erreurs2} shows the rotation and translation errors as a function of Gaussian noise added to the rotational part of the hand and camera positions. These errors are obtained with the three methods. We can conclude that the closed-form method is more accurate than the linear method proposed in \cite{Zhuang94}.

As other authors have done in the past, it is interesting to analyze the
behaviour of calibration methods with respect to the number of positions. In
order to perform this analysis we have to fix the percentage of noise. Figure~\ref{fig:erreurs4} shows the rotational
and translational errors as a function of the square root of the number of
motions ($\sqrt{n}$ varies from $1.732$ to $2.828$). The noise ratio has been
fixed to the worst case for rotations, e.g., 6\% and to 2\% for translations. Both rotational and translational noise distributions are Gaussian. 

\section{Experimental results}
\label{sec:experimental}

In this Section we report some experimental results obtained with two sets of
data. 
The first data set was obtained with 17 different positions of the hand-eye device with respect to a
calibrating object. The second data set was obtained with 7 such positions.
In order to calibrate the camera we used the classical
method proposed by Faugeras \&
Toscani described in \cite{Faugeras93}. 

Our tests compare the linear method \cite{Zhuang94} with the two methods developed in this paper.
Table~\ref{table-1} and Table~\ref{table-2} summarize the results obtained with the two data sets mentioned above. 
The second columns of these tables show the sum of squares of the absolute error in rotation. The third columns show the relative error in translation, namely
\begin{eqnarray}
\label{eq:ER}
E_{\Rmat} &=& \sum \| \Rmat_{A}\Rmat_{X}-\Rmat_{Z}\Rmat_{B} \|^{2} \\
\label{eq:Et}
E_{\tvect} &=& \left ( \frac{\sum \| (\Rmat_{A} t_{X}+ t_{A} - \Rmat_{Z} t_{B} - t_{Z}  \|^{2}}
      {\sum \| \Rmat_{A} t_{X} + t_{A} \|^{2}} \right ) ^{1/2} 
\end{eqnarray}

\begin{table}[h!]
\caption{The formulation $\Amat\Xmat=\Zmat\Bmat$ used with the first data set (17 different positions of the hand-eye device). These data were obtained with a PPPRRR robot.}
\begin{center}
\begin{tabular}{|c||c|c|}
\hline
         & $E_{\Rmat} $, eq. (\ref{eq:ER}) &
$E_{\tvect} $, eq. (\ref{eq:Et})   \\
\hline
Linear solution    & 0.00031   & 0.00068 \\
\hline 
Closed-form solution &   0.00026        & 0.00075          \\
\hline
Non-linear optimization & 0.00071   & 0.00021 \\
\hline
\end{tabular}
\end{center}
\label{table-1}
\end{table}
\begin{table}[h!]
\caption{The formulation $\Amat\Xmat=\Zmat\Bmat$ used with the second data set (7 different positions of the hand-eye device). These data were obtained with a RRRRRR robot.}
\begin{center}
\begin{tabular}{|c||c|c|}
\hline
         & $E_{\Rmat} $, eq. (\ref{eq:ER}) &
$E_{\tvect} $, eq. (\ref{eq:Et})  \\
\hline
Linear solution   & 0.12174  & 0.00738 \\
\hline 
Closed-form solution &  0.00068        &   0.00515     \\
\hline
Non-linear optimization & 0.00109   & 0.00451 \\
\hline
\end{tabular}
\end{center}
\label{table-2}
\end{table}

It is worthwhile to notice that the robots being used in the two experiments
summarized in the tables above are not identical. The first data set
(Table~\ref{table-1}) was obtained with a PPPRRR robot (three prismatic and
three rotational joints) while the second data set (Table~\ref{table-2}) was
obtained with a RRRRRR robot.
Unlike the simulated data, these two experiments do not allow one to conclude
that the closed-form solution outperforms the linear solution. In the first
experiment the linear solution yields a smaller translation error 
than the translation error associated with the closed-form method. In the
second experiment the translation error associated with the linear method does
not seem to be affected by a large rotation error.

These experimental results seem however to confirm that the non-linear method provides a better estimation of the translation vectors {\em at the cost of slightly larger rotation errors}. This is due to the fact that the robot's kinematic chain is not perfectly calibrated and therefore there are errors associated with the robot's translation parameters. Obviously, these errors do not obey the noise models used for simulations. 

\section{Discussion}
\label{sec:discussion}
In this paper we addressed the problem of robot-to-world and hand-eye calibration. As it was proposed in \cite{Zhuang94}
this problem is formulated as solving a system of homogeneous transformation equations of the form $\Amat\Xmat=\Zmat\Bmat$. 

We develop two resolution methods, the first one solves for rotations and then for translations
while the second one solves simultaneously for rotations and translations. The first method leads to a closed-form solution while the second one leads to non-linear optimization. 

Both the sensitivity analysis and the results obtained with experimental data
show that the closed-form method slightly outperforms the linear method
of Zhuang et al. \cite{Zhuang94}. This is most probably due to the Euclidean
nature of the error function suggested in Section~\ref{section:closed-form-solution}. However, there is no evidence that with real data the
closed-form method will always perform better than the linear method:
One can therefore conclude that the two methods have comparable performances.

The non-linear minimization method suggested in Section~\ref{section:non-linear-solution} 
yields the
most accurate results and outperforms both the linear and closed-form methods. 
The solution obtained with either the linear or closed-form methods can be
used to initialize the non-linear minimization method.

The two methods proposed in this paper together with \cite{Zhuang94}
may be useful to other problems that can be formulated into homogeneous transformation equations of the form $\Amat\Xmat=\Zmat\Bmat$. 

\balance

\begin{thebibliography}{10}
\providecommand{\url}[1]{#1}
\csname url@samestyle\endcsname
\providecommand{\newblock}{\relax}
\providecommand{\bibinfo}[2]{#2}
\providecommand{\BIBentrySTDinterwordspacing}{\spaceskip=0pt\relax}
\providecommand{\BIBentryALTinterwordstretchfactor}{4}
\providecommand{\BIBentryALTinterwordspacing}{\spaceskip=\fontdimen2\font plus
\BIBentryALTinterwordstretchfactor\fontdimen3\font minus
  \fontdimen4\font\relax}
\providecommand{\BIBforeignlanguage}[2]{{%
\expandafter\ifx\csname l@#1\endcsname\relax
\typeout{** WARNING: IEEEtran.bst: No hyphenation pattern has been}%
\typeout{** loaded for the language `#1'. Using the pattern for}%
\typeout{** the default language instead.}%
\else
\language=\csname l@#1\endcsname
\fi
#2}}
\providecommand{\BIBdecl}{\relax}
\BIBdecl

\bibitem{Zhuang94}
H.~Zhuang, Z.~S. Roth, and R.~Sudhakar, ``{Simultaneous robot/world and
  tool/flange calibration by solving homogeneous transformation equations of
  the form AX= YB},'' \emph{IEEE Transactions on Robotics and Automation},
  vol.~10, no.~4, pp. 549--554, 1994.

\bibitem{TsaiLenz89}
R.~Tsai and R.~Lenz, ``A new technique for fully autonomous and efficient 3{D}
  robotics hand/eye calibration,'' \emph{IEEE Journal of Robotics and
  Automation}, vol.~5, no.~3, pp. 345--358, June 1989.

\bibitem{Wang92}
C.-C. Wang, ``Extrinsic calibration of a robot sensor mounted on a robot,''
  \emph{IEEE Transactions on Robotics and Automation}, vol.~8, no.~2, pp.
  161--175, April 1992.

\bibitem{Park94}
F.~C. Park and B.~J. Martin, ``{Robot sensor calibration: solving AX= XB on the
  Euclidean group},'' \emph{IEEE Transactions on Robotics and Automation},
  vol.~10, no.~5, pp. 717--721, 1994.

\bibitem{ShiuAhmad89}
Y.~C. Shiu and S.~Ahmad, ``Calibration of wrist mounted robotic sensors by
  solving homogeneous transform equations of the form {AX=XB},'' \emph{IEEE
  Journal of Robotics and Automation}, vol.~5, no.~1, pp. 16--29, February
  1989.

\bibitem{HoraudDornaika95}
R.~Horaud and F.~Dornaika, ``Hand-eye calibration,'' \emph{International
  Journal of Robotics Research}, vol.~14, no.~3, pp. 195--210, June 1995.

\bibitem{ZhuangWangRoth95}
H.~Zhuang, K.~Wang, and Z.~S. Roth, ``Simultaneous calibration of a robot and a
  hand-mounted camera,'' \emph{IEEE Transactions on Robotics and Automation},
  vol.~11, no.~5, pp. 649--660, October 1995.

\bibitem{WalkerShaoVolz92}
M.~W. Walker, L.~Shao, and R.~A. Volz, ``Estimating 3-d location parameters
  using dual number quaternions,'' \emph{CGVIP-Image Understanding}, vol.~54,
  no.~3, pp. 358--367, November 1991.

\bibitem{Faugeras93}
O.~D. Faugeras, \emph{Three Dimensional Computer Vision: A Geometric
  Viewpoint}.\hskip 1em plus 0.5em minus 0.4em\relax Boston: MIT Press, 1993.

\bibitem{GillMurrayWright89}
P.~E. Gill, W.~Murray, and M.~H. Wright, \emph{Practical Optimization}.\hskip
  1em plus 0.5em minus 0.4em\relax London: Academic Press, 1989.

\bibitem{Fletcher90}
R.~Fletcher, \emph{Practical Methods of Optimization}.\hskip 1em plus 0.5em
  minus 0.4em\relax John Wiley \& Sons, 1990.

\bibitem{NumericalRecipes}
W.~Press, B.~Flannery, S.~Teukolsky, and W.~Wetterling, \emph{Numerical Recipes
  in C: The Art of Scientific Computing}.\hskip 1em plus 0.5em minus
  0.4em\relax Cambridge University Press, 1988.

\end{thebibliography}

\end{document}